\documentclass[10pt,twocolumn,letterpaper]{article}

\usepackage{cvpr}
\usepackage{times}
\usepackage{epsfig}
\usepackage{graphicx}
\usepackage{amsmath}
\usepackage{amssymb}
\usepackage{bm}
\usepackage{tocloft}
\usepackage{float}
\usepackage[pagebackref=true,breaklinks=true,letterpaper=true,colorlinks,bookmarks=false]{hyperref}

\usepackage[breaklinks=true,bookmarks=false]{hyperref}
\usepackage[title]{appendix}

\cvprfinalcopy 

\def\cvprPaperID{****} 

\usepackage{subcaption}
\DeclareCaptionSubType*[Alph]{table}
\DeclareCaptionLabelFormat{mystyle}{Table~\bothIfFirst{#1}{ }#2}
\usepackage{bm}
\usepackage[pagebackref=true,breaklinks=true,letterpaper=true,colorlinks,bookmarks=false]{hyperref}

\ifcvprfinal\fi
\begin{document}
\title{Adversarial Defense by Stratified Convolutional Sparse Coding}
\author{Bo Sun\\
Peking University\\
\tt\small bosun@pku.edu.cn
\and
Nian-hsuan Tsai\\
National Tsinghua University\\
\tt\small nianhsuan@gmail.com
\and
Fangchen Liu \quad Ronald Yu \quad Hao Su\\
UC San Diego\\
\tt\small \{fliu,ronaldyu,haosu\}@eng.ucsd.edu
}

\maketitle

\begin{abstract}
 \vspace{-3mm} 

We propose an adversarial defense method that achieves state-of-the-art performance among attack-agnostic adversarial defense methods while also maintaining robustness to input resolution, scale of adversarial perturbation, and scale of dataset size.
Based on convolutional sparse coding, we construct a stratified low-dimensional quasi-natural image space that faithfully approximates the natural image space while also removing adversarial perturbations. We introduce a novel  Sparse Transformation Layer (STL) between the input image and the first layer of the neural network to efficiently project images into our quasi-natural image space.
Our experiments show state-of-the-art performance of our method compared to other attack-agnostic adversarial defense methods in various adversarial settings. 
\end{abstract}

 \vspace{-3mm}

\section{Introduction}

Existing defense mechanisms against adversarial attacks, although able to achieve robustness in certain adversarial settings, are still unable to achieve true robustness to all adversarial inputs. The most effective existing defense methods modify the network training process to improve robustness against adversarial examples~\cite{Goodfellow14fgsm, Kurakin16ifgsm, tramer2018ensemble,Na18cascadeadvtrain}. However, they are trained to defend a specified attack for a  specified model, limiting their real-world applications and claims of robustness to \textit{all} adversarial inputs. Ideally, our defense mechanism should be \textit{attack agnostic} and \textit{model agnostic}. 

Instead of modifying the network and training process, another line of existing methods achieve the desired property of being attack-agnostic and model-agnostic by modifying adversarial inputs to resemble clean inputs~\cite{Das17jpeg, Dziugaite16jpeg, Osadchy17filters,xie2018mitigating, Prakash18deflect,song18pixeldefend, Meng17magnet,Samangouei18defensegan}. 
However, these methods show weaknesses in other adversarial settings such as being  unable to handle larger perturbations, unable to simultaneously handle many different resolutions, and not scalable to large datasets.

\begin{figure}
	\centering
	\includegraphics[width=1\linewidth]{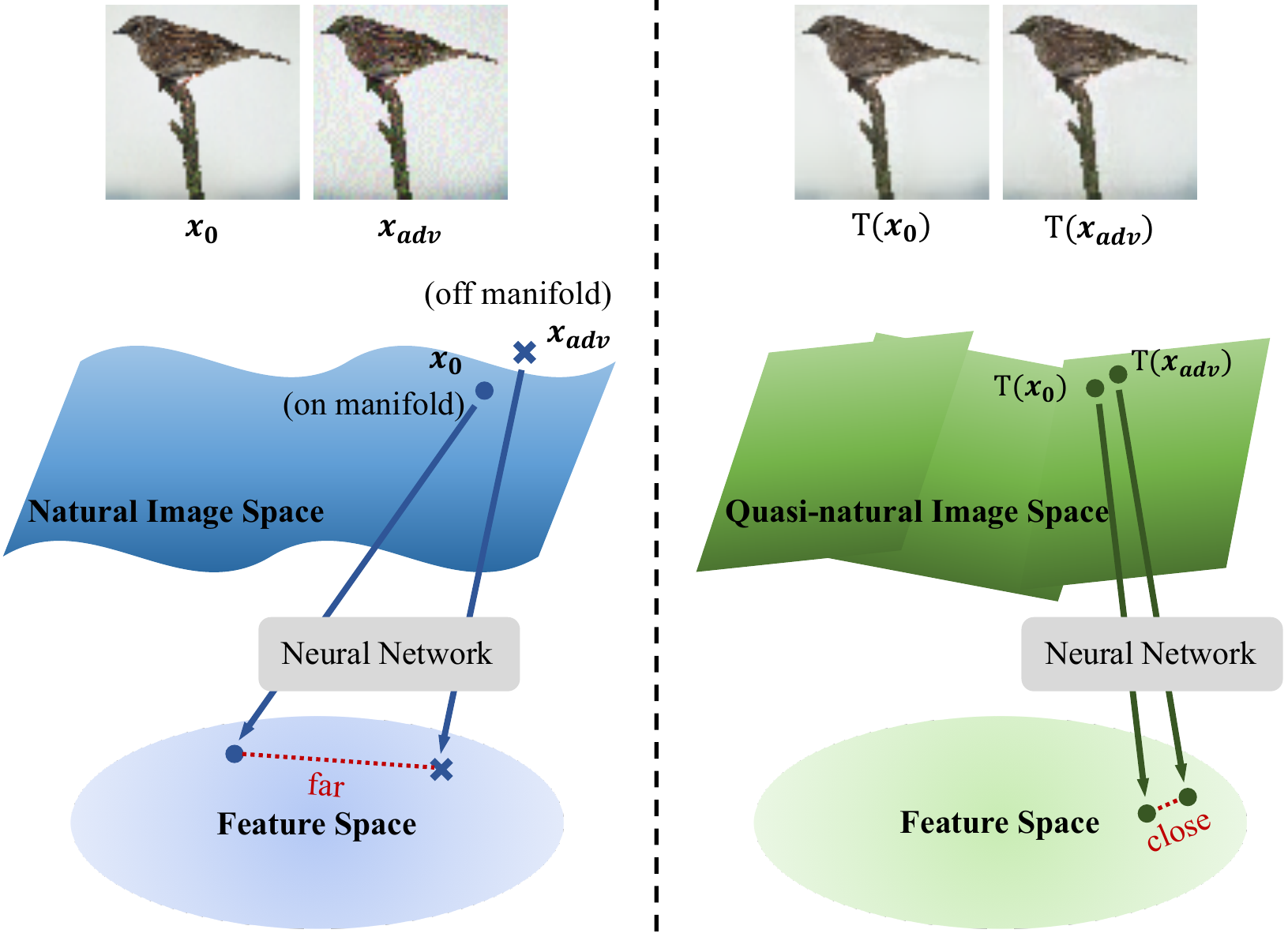}
	\caption{Comparison of feature extraction between natural image space and our learned quasi-natural image space. In the natural image space, a neural network trained from natural images may assign different labels to the adversarial example and the clean image, since they can be far from each other in the feature space. After projection to the quasi-natural image space, they tend to lie closely together in the feature space.}
	\vspace{-3mm}
	\label{fig::overview}
\end{figure}

In this paper, we present an input-transformation based defense method that
achieves state-of-the-art performance  when compared to previous attack-agnostic and model-agnostic defense methods.
Moreover, our method is also far more simultaneously robust to scale of attack perturbation, a variety of different input resolutions, and dataset scale.
We achieve our high level of robustness by projecting both clean and adversarially attacked input images into a low-dimensional \textit{quasi-natural image space} that faithfully approximates the natural image space while also removing adversarial perturbations so that adversarial examples will be close to their original inputs in feature space. 

We construct the quasi-natural image space in an unsupervised manner using a convolutional dictionary learning-based~\cite{Bristow13fastcsc, Heideffcsc, Chalasani13fastproxcsc,Wohlberg14efficientcsc} method, and we project the input images into our quasi-natural image space by introducing a novel \textit{Sparse Transformation Layer (STL)} between the input and first layer of the network. 
We can further enhance the robustness of our pipeline by  retraining a classifier on the quasi-natural images.

Experimentally, we demonstrate that our method achieves a significant robustness improvement in a variety of different adversarial settings compared with state-of-the-art attack-agnostic defense methods. 
We also show that our quasi-natural image space is able to provide a better blend of preservation of image details and ability to remove adversarial perturbations compared to other input-transformation-based adversarial defense methods. 

In summary, our contributions are:
 \vspace{-1mm} 
\begin{itemize}
	\item We propose a novel and effective attack-agnostic adversarial defensive method that uses a novel \textit{Sparse Transformer Layer} to transform images so that  corresponding clean and adversarial images lie close together both in our \textit{quasi-natural image space} and feature space. 
	 \vspace{-1mm} 
	\item We demonstrate that our defense method achieves state-of-the-art performance among attack-agnostic adversarial defense methods.
	 \vspace{-1mm} 
	\item Compared to previous state-of-the-art, our defense method is far more capable of effectively handling a variety of image resolutions, large and small image perturbations, and large-scaled datasets.
	 \vspace{-1mm} 
	\item Among image-transformation-based adversarial defenses, our image projection onto quasi-natural image space achieves the best blend of image detail preservation and ability to remove adversarial perturbations.
\end{itemize}
\section{Related Works}
\label{sec::related}
\paragraph{Adversarial Attacks}
Adversarial attacks are inputs that are intentionally slightly perturbed to fool machine learning models. Szegedy et al.~\cite{szegedy14intrigue} first introduce adversarial examples and generate them with the box-constrained L-BFGS method. Goodfellow et al.~\cite{Goodfellow14fgsm} propose an efficient single step attack called FGSM based on network linearity. Kurakin et al.~\cite{Kurakin16bim} apply FGSM iteratively and propose BIM. DeepFool~\cite{deepfool} finds the smallest perturbation crossing the model decision boundary. CW~\cite{Carlini17cw} solves an optimization problem which minimizes both the objective function and difference between adversarial and clean images. Liu et al.~\cite{Liu16transfer} generate strongly transferable adversarial examples with an ensemble-based approach. Non-gradient based attacks such as one pixel attack~\cite{su17onepixel} and Zoo~\cite{chen17zoo} do not require knowledge of network parameters and architecture. 
 \vspace{-3mm} 
\paragraph{Adversarial Defense via Network Modification} This type of defense aims to improve the robustness of the target model against adversarial examples. The most common method is adversarial training~\cite{Goodfellow14fgsm,Kurakin16ifgsm, tramer2018ensemble, Na18cascadeadvtrain} which adds adversarial examples into training data. This class of methods effectively enhances robustness to the adversarially trained attacks but has poor generalizability to unknown attacks. Other methods like feature squeezing~\cite{Xu18featuresqueeze}, network distillation~\cite{Papernot16distillation}, region-based classifier~\cite{cao17region} and saturating networks~\cite{Nayebi17saturating} modify the learning strategy based on gradient masking~\cite{Papernot16masking} and smooth the decision boundary, but they are still vulnerable to black-box attacks~\cite{Carlini17cw, Papernot17black}. 
 \vspace{-3mm} 
\paragraph{Adversarial Defense via Input Transformation} Input-transformation defenses aim to remove adversarial perturbation transforming inputs before feeding them to the target network.  Some previous methods treat adversarial perturbation as high frequency noise and resort to traditional denoising methods to smooth small perturbations. ~\cite{Das17jpeg, Dziugaite16jpeg} study the effect of JPEG compression on removing adversarial noise. Osadchy et al.~\cite{Osadchy17filters} apply a set of filters such as median filter and averaging filter to remove perturbation. Guo et al.~\cite{Guo18inputtrans} test five transformations and find total variation minimization and image quilting obtain good defensive performance. These denoising methods only fix small perturbations and  suffer from information loss. 

More recently, other works have tried to purify adversarial images through generative models. Meng et al.~\cite{Meng17magnet} propose a two-pronged defense mechanism and use a denoising auto-encoder to remove adversarial perturbation on MNIST digits~\cite{LeCun98mnnist}. Song et al.~\cite{song18pixeldefend} transform adversarial images into clean images using PixelCNN~\cite{Salimans17pixelcnn}. Although they achieve good performance on small datasets, these methods do not scale well to higher-resolution or larger datasets.

Pixel manipulation methods are also used to remove small adversarial perturbations. Xie et al.~\cite{xie2018mitigating} utilize random resizing and padding to mitigate adversarial effects. Prakash et al.~\cite{Prakash18deflect} locally corrupt adversarial images by redistributing pixel values via a process we term pixel deflection. However, these methods suffer when they encounter perturbations that are not extremely small. 

Most similar to our method is D3~\cite{Dezfooli18patch}, which denoises adversarial images by replacing patches with a sparse combination of natural images patches. Further discussion of D3 is reserved for Section~\ref{sec::analysis}.
\vspace{-3mm} 
\paragraph{Convolutional Dictionary Learning}
Convolutional sparse representations are a form of sparse representation learning~\cite{Mairal14sparserepre} with a dictionary that has a structure that is equivalent to convolution with a set of linear filters~\cite{Cardona17cdlreview,Bristow13fastcsc}. It is widely and successfully used in signal processing and computational imaging~\cite{Gu15cdlsuperresolutioin,Liu16cdlfusion,Quan16cdlconstrast,Zhang18cdldecomposition,Zhang17cdlstreak,Serrano16cdlhdr}. Many efficient algorithms~\cite{Heideffcsc, Bristow13fastcsc, Chalasani13fastproxcsc, Wohlberg14efficientcsc,Choudhury17consensus} have been developed to solve this problems. 
Sung et al. also recently introduced a method that used a deep neural network to learn sparse dictionaries for 3D point clouds ~\cite{sung2018deep}.

\section{Approach}
\subsection{Method Overview}
\label{sec::overview}
Let $\mathcal{X}$ be the image space and $\mathcal{Y}$ be the label space. $f_\theta(\cdot): \mathcal{X} \rightarrow \mathcal{Y}$ is a classifier parameterized by $\theta$. Given the classifier $f_\theta(\cdot)$ and a clean image $\bm{x_0}$, an adversarial example $\bm{x}_{adv}=\bm{x_0} + \bm{\eta}$ is an image slightly different from $\bm{x_0}$ but confuses $f$:
\begin{equation}
d(\bm{x}_{adv}, \bm{x_0})<\epsilon \mbox{ but } f_\theta (\bm{x}_{adv}) \neq f_\theta (\bm{x_0}), 
\end{equation}
where $d(\cdot, \cdot)$ is a distance function between the clean and adversarial images. 
$\epsilon$ is the perturbation scale which is often set to a small number to get almost imperceptible difference between $\bm{x}_{adv}$ and $\bm{x_0}$. 

Adversarial examples $\bm{x}_{adv}$ are fabricated images and usually lie out of the natural image manifold. This may cause the network trained from natural images, even with adversarial data augmentation, to map $\bm{x}_{adv}$ far away from $\bm{x_0}$ (Figure~\ref{fig::overview} Left). Our idea is thus to recover $\bm{x_0}$ as much as possible by projecting $\bm{x}_{adv}$ to the natural image manifold. However, parameterizing the true natural image manifold is practically infeasible. We instead leverage manifold learning to build a low-dimensional space that approximates the natural image space, which we dub the \emph{quasi-natural space} $\mathcal{P}$ in this paper. Along with $\mathcal{P}$, there is a transformation $T$ that maps an image (natural or spoofed) to $\mathcal{P}$. We require that $T$ satisfies the following constraints: 

\begin{enumerate}
    \item $f_\theta (T(\bm{x}_{adv})) = f_\theta (T(\bm{x_0})) = y_{\bm{x_0}}$;
    \item $d(T(\bm{x}_{adv}), T(\bm{x_0}))\ll\epsilon$.
\end{enumerate}

Condition 1 requires that the classifier $f$ assigns the same groundtruth label to $\bm{x}_{adv}$ and $\bm{x_0}$, which is our final goal. To guarantee Condition 1, other than learning $f$ to optimize classification accuracy, we also introduce Condition 2 (Figure~\ref{fig::overview} Right). Condition 2 requires that $\bm{x}_{adv}$ and $\bm{x_0}$ should be situated closely in $\mathcal{P}$, so that we can learn a quite smooth function $f$ satisfying Condition 1. This is important since our $f$ is a neural network, and learning a smoother map would endow the it better generalization power. 

We take an unsupervised approach to build the quasi-natural image space. This space is constructed by stitching multiple low-dimensional linear subspaces together. Practically, we cluster the training data into a few groups and we learn a linear subspace for each group by convolutional sparse coding algorithm~\cite{Heideffcsc,Choudhury17consensus}. With this quasi-natural space constructed, we are able to project any image to this space by the sparse transformation layer introduced in Section~\ref{sec::stl}, which will remove a significant amount of adversarial perturbations. Then in this quasi-natural image space we can retrain a classifier to allow robust prediction over adversarial examples (Section~\ref{sec::traincnn}).


\subsection{Sparse Transformation Layer (STL)} 
\label{sec::stl}
\begin{figure*}
	\centering
	\includegraphics[width=1\linewidth]{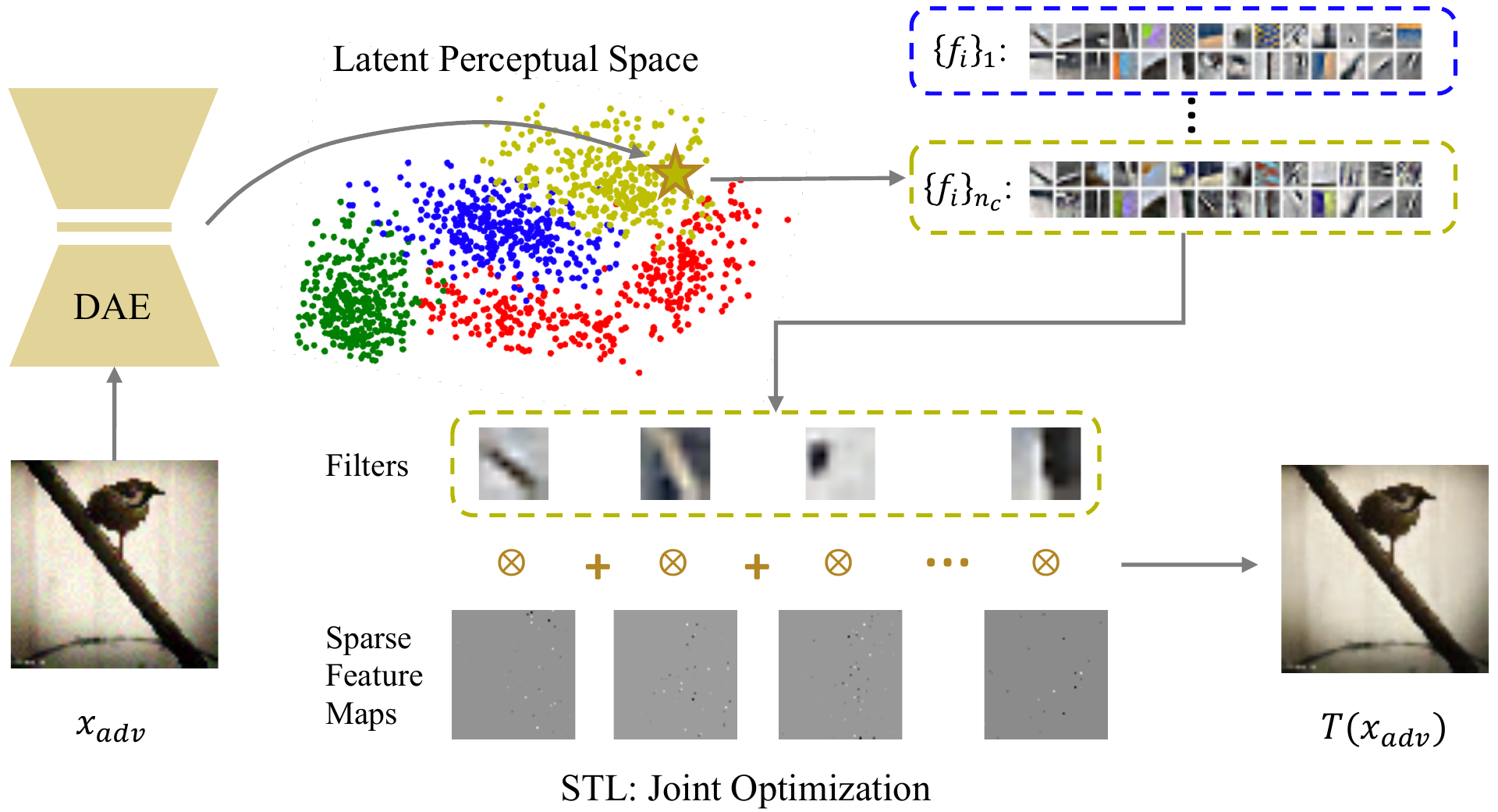}
	\vspace{-2mm}
	\caption{Pipeline of our defensive method. We first feed an image to a pre-trained Denoising Auto-Encoder and find the cluster the image should belong to. Then we select the dictionary corresponding to the selected cluster and jointly optimize the sparse feature maps and filters in this dictionary. In this way, we can project the input to the quasi-natural image space.}
	\vspace{-2mm}
	\label{fig::pipeline}
\end{figure*}

Given a classification network $f$, we add a Sparse Transformation Layer (STL) between the input image and the first layer of $f$. This STL layer projects the input (adversarial or clean) onto a quasi-natural space, which removes nuisances including adversarial perturbations in the appearance. 

Let the projection of $\bm{x}$ be $T(\bm{x})$ (assume that $\bm{x}$ is an image of $C$ channels). The projection in our STL layer follows from the Convolutional Sparse Coding algorithm~\cite{Choudhury17consensus}. This algorithm learns a dictionary in a convolutional manner by solving the following optimization problem: 
\begin{equation}
\begin{aligned}
 \underset{\{\bm{f}_{i,c}\}, \{\bm{z}_i\}}{\mbox{minimize}} & &\frac{1}{2} \sum_{c=1}^{C}\|\bm{x}_c-T(x)_c\|_2^2 + \lambda \sum_{i=1}^{K} \|\bm{z}_i\|_1\\
 \mbox{subject to} & & T(x)_c=\sum_{i=1}^{K} \bm{f}_{i,c} \otimes \bm{z}_i\\
 & & \|\bm{f}_{i,c}\|_2^2=1, 1\le i \le K,1\le c\le C
\end{aligned}
\label{prob1}
\end{equation}
where $\otimes$ indicates the convolution operator, $C$ is the number of input channels, $K$ is the number of filters for each input channel, ${\bm{f}_{i, c}}|_{i={1,...,K}; c={1,...,C}}$ denotes a set of filters, and ${\bm{z}_i}|_{i={1,...,K}}$ are the feature maps for each filter. 

Different from standard sparse coding, which learns a dictionary and code for the whole image, as shown by ~\cite{Tosic11diclearn}, Problem \eqref{prob1} learns to reconstruct image patches by local dictionaries and codes. Here, the local dictionary contains the set of filters $\bm{f}_{i,c}$, and local codes are stored in the feature map $\bm{z_i}$. The convolution operation in the constraint essentially computes the linear combination of local filters. In vanilla sparse coding, a small set of bases are selected to reconstruct the image. Similarly, in the convolutional sparse coding formulation, a small set of filters should be selected to reconstruct a local patch. To achieve the filter selection goal, we have to enforce the feature map $\bm{z_i}$ to be sparse by adding the $\ell_1$ regularization term.

In practice, we prefer to use a small number of filters. This forces filters to learn major and expressive local patterns on the natural image manifold. Moreover, from our observation, having too many filters may cause extra filters to learn high frequency components, which can be used to reconstruct arbitrary image patches including adversarial perturbation that should be removed. 
\vspace{-1mm}

\subsection{Learning Filters and Feature Maps}
\label{sec::learn}
Plugging the constraint in Problem~\eqref{prob1} into the objective function, we see that Problem \eqref{prob1} is biconvex in $\bm{f}_{i,c}$ and $\bm{z_i}$. To solve this biconvex problem, we alternate between (1) learning shared filters from clean images, and (2) learning sparse feature maps for each input image with fixed filters. Next we briefly introduce these two stages.
\vspace{-4mm}

\paragraph{Dictionary Learning.} 
Given feature maps, Problem~\eqref{prob1} becomes convex in $\bm{f}_{i, c}$. To solve this problem efficiently, we transform to the Fourier domain~\cite{Wohlberg2014efficientCS} and use ADMM algorithm as the solver following the framework of \cite{Choudhury17consensus}.
\vspace{-4mm}
\paragraph{Sparse feature map (code) learning.}
Given fixed filters $\{\bm{f}_{i,c}\}$, our objective function is again a convex optimization problem in $\bm{z_i}$.
The problem is also known as Convolutional Basis Pursuit DeNoising (CBPDN)~\cite{Chen98bpdn} and we use ADMM algorithm~\cite{Boyd10admm} to solve it.

\subsection{Stratified Quasi-Natural Image Space }
\label{sec::cluster}

Due to the high inherent variation of natural images, it is hard to well reconstruct all images using just a small dictionary. However, as we discussed at the end of Section~\ref{sec::stl}, we also do not want to employ a too big dictionary, because the big dictionary will span an excessively high-dimensional space, inevitably covering a significant amount of non-natural images. This would reduce the power of our algorithm to filter out adversarial perturbations. 

To circumvent the challenge, we split the data manifold into several regions and learn an individual small dictionary for each region. In this way, each image is still reconstructed by a small dictionary, but we can still reconstruct all images well using their corresponding dictionaries. 

In practice, we partition the image space by clustering natural image samples based on their perceptual features.
Generative models can learn perceptual features by reconstruction loss. In particular, we find that Denoising Auto-Encoder (DAE)~\cite{Bengio13dae} fits the adversarial setting well because it is trained with noisy input and the feature extraction process can modestly tolerate input noise. Specifically, we train a DAE on both natural images and their noise-perturbed versions (Gaussian noise). In practice we find that the original image and adversarial attacked version usually live closely in the latent space learned by DAE. We then use the K-means algorithm to cluster training data~\cite{David07kmeans}. 

The clusters allow us to partition the natural image manifold. Given an arbitrary input image (adversarial or clean), we can obtain its latent features from the DAE and find the $k$-nearest neighbors in the training image dataset. Then we vote for the cluster the image should belong to. Once we have found the cluster, we can either update the filters and features maps for dictionary learning, or compute the projection of the image for classification network training/test.

\subsection{Classifier Training in the Quasi-Natural Space}
\label{sec::traincnn}
To train a classifier for image categorization, we map all the clean training images to $\mathcal{P}$.  We simply use their reconstructed version $T(\bm{x_0})$ to train a user-selected classification network (e.g., AlexNet). To perform defense at test time, we apply the trained classifier on the $T$-transformed version of the testing image (clean or adversarial).

After projection to $\mathcal{P}$, $T(\bm{x}_{adv})$ and $T(\bm{x_0})$ share close perceptual and semantic features. Therefore, decisions made in this quasi-natural space $\mathcal{P}$ tend to be more reliable for adversarial examples compared to the original space.   


\section{Discussion}
\label{sec::analysis}
In this section, we discuss our unique advantages over existing adversarial defenses and then analyze possible reasons behind the effectiveness of our method against popular gradient-based attack methods. 
 \vspace{-3mm} 
\paragraph{Relationship With Existing Methods}
In contrast to adversarial learning methods~\cite{Goodfellow14fgsm,Kurakin16ifgsm,tramer2018ensemble} that rely on direct knowledge of the attack method and model type, our algorithm only relies on the clean training data at hand. Built without any explicit prior knowledge of the attacker, our design does not overfit to any specific attack strategy and tends to be a generic tool. 


Recent attack-agnostic defense methods use generative models to transform images into a low-dimensional space~\cite{Meng17magnet,song18pixeldefend, Samangouei18defensegan}. We choose not to use a network to build our low-dimensional space, since the generative network itself is vulnerable to adversarial attacks. Another disadvantage of these methods is that the limited expressive power of generative models restricts the domain of these methods to datasets small in resolution and scale such as MNIST~\cite{LeCun98mnnist} and CIFAR-10~\cite{cifar}. Pixel manipulation methods~\cite{Prakash18deflect, xie2018mitigating} can work on large datasets, but they only achieve good performance under extremely small perturbations. Our method works uniquely well on large adversarial perturbation, complicated datasets, and higher resolutions. 

The D3 algorithm proposed in \cite{Dezfooli18patch} is the most similar to ours. It replaces noisy adversarial image patches by a sparse combination of natural image patches. However, our method provides several advantages. First, D3 reconstructs images poorly on low-resolution datasets like CIFAR-10~\cite{cifar}. Second, the size of the natural patch dictionary is very large (10K-40K) while we only need a small number of filters (typically 64). The large size of their patch dictionary has two main drawbacks: the excessive number of dictionary elements may lead the dictionary to learn high frequency components, which can be used to wrongly reconstruct adversarial perturbations, and the generic dictionary elements are not as expressive as ours, so D3 generates images that are not as sharp as ours as verified in our experiments.

\vspace{-3mm}
\paragraph{Robustness to Gradient-Based Attacks}

There are two main concepts behind the effectiveness of our method against gradient-based attacks: (1) \emph{Gradient Obfuscation:} 
Obtaining the numerical gradient of the STL is likely to be challenging, because the output of the STL is the solution to a non-convex optimization problem (has the $\arg\min$ form of the input image). Without the gradient of the STL, designing gradient-based attack becomes difficult. (2) \emph{High-frequency Perturbation Removal:} Existing gradient-based attack mechanisms often introduce high-frequency perturbations. With a small dictionary and the sparsity constraint in Problem~\eqref{prob1}, the learned filters tend to be quite smooth (Figure~\ref{fig::pipeline}), which could filter out the high-frequency perturbation patterns.
\section{Experiments}
\begin{figure*}
	\centering
	\includegraphics[width=0.9\linewidth]{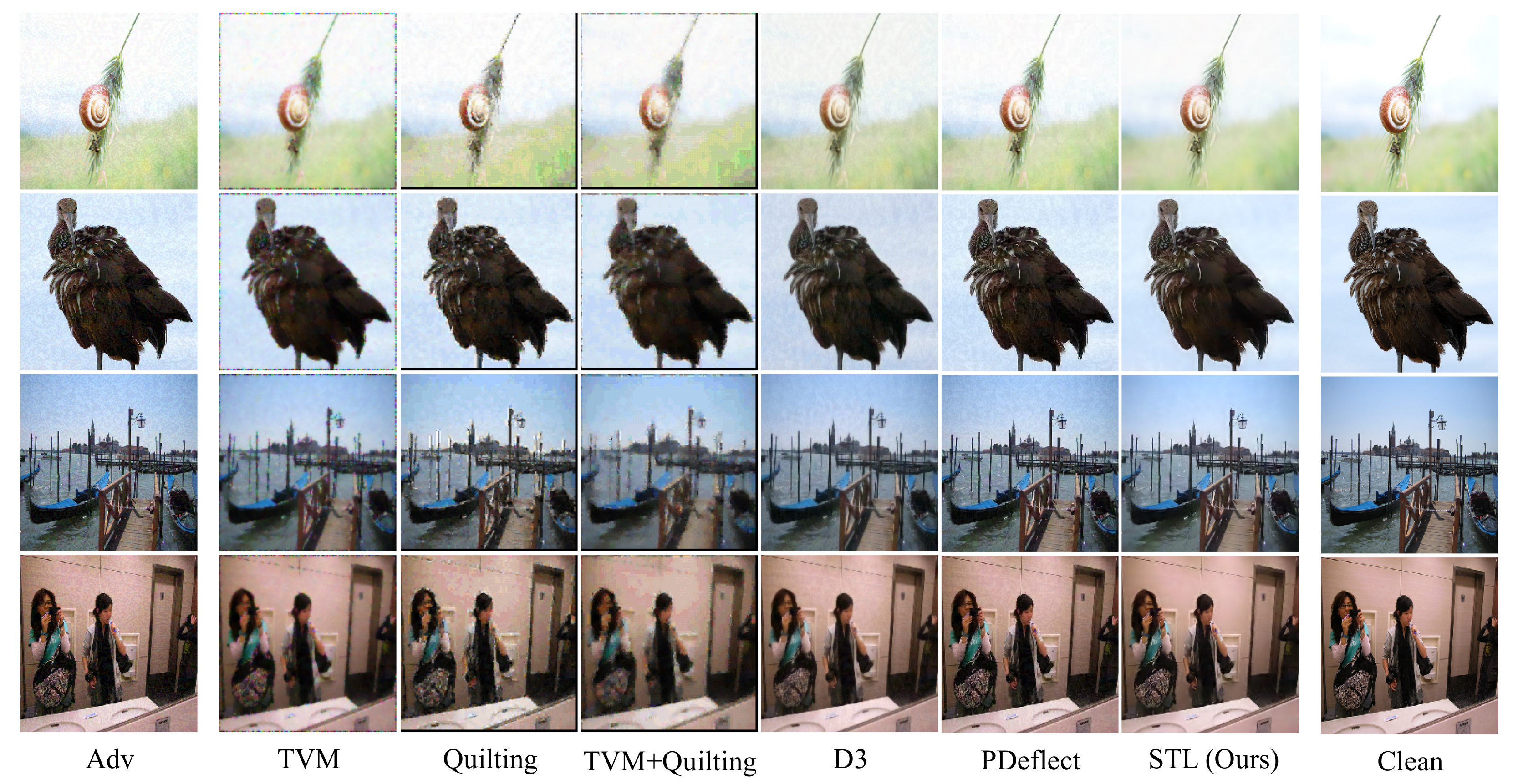}
	\caption{Qualitative comparison of image reconstruction results on ImageNet. The first column is input the adversarial examples generated by FGSM~\cite{Goodfellow14fgsm} attack with $L_2$ dissimilarity $0.08$. The last column is the corresponding clean images. Visually, our method outperforms others on removing adversarial perturbations and keeping input details. D3 refers to \cite{Dezfooli18patch} and PDeflect refers to \cite{Prakash18deflect}. }
	\label{fig::comparison_imagenet}
\end{figure*}

In this section, we first introduce our experimental settings, and then show a quantitative and qualitative comparison with other attack-agnostic adversarial defenses. We demonstrate that our method outperforms the state-of-the-art. Lastly, we perform an analysis of the intrinsic trade-off between projection image quality and defense robustness of transformation-based defenses. 

\subsection{Settings}

We conduct experiments on CIFAR-10~\cite{cifar}, ImageNet~\cite{Deng09imagenet}, and ImageNet-10, where we manually choose 10 coarse-grained classes from the whole dataset, e.g. bird, car, cat, etc. Every class contains 8000 training and 2000 testing images. 




We evaluate our method on VGG-16~\cite{Simonyan14vgg} and ResNet-50~\cite{He17resnet} to defend against   FGSM~\cite{Goodfellow14fgsm}, BIM~\cite{Kurakin16bim}, DeepFool~\cite{deepfool}, and CW~\cite{Carlini17cw}. We constrain the perturbation scale  $\|\bm{\eta}\|_2=\frac{\|\bm{x}_{adv}-\bm{x_0}\|}{\|\bm{x_0}\|}$ to 0.04 (FGSM-0.04) and 0.08  (FGSM-0.08) for FGSM and to 0.04 for BIM, DeepFool, and CW. 

\label{sec::attacksetting}

By default, we set the filter number $K=64$, filter size $S=8$, and sparse constraint $\lambda=0.2$. We first downsample images to $32\times 32$ to train a DAE, and split the latent space to 4 clusters for CIFAR-10 and ImageNet-10, and 10 clusters for ImageNet.

\subsection{Adversarial Defense} 
We evaluate the defensive effectiveness of our method of retraining a classifier on quasi-natural images and then projecting adversarial examples onto the quasi-natural image space as described in Section ~\ref{sec::traincnn}.

 Classification accuracy comparison results are in Table ~\ref{tab::cifar-proj} for CIFAR-10, Table ~\ref{tab::imagenet10-proj} for ImageNet-10 and Table ~\ref{tab::imagenet-proj} for ImageNet.
  In Table ~\ref{tab::cifar-proj} and Table ~\ref{tab::imagenet10-proj} we follow our setting as described in Section \ref{sec::attacksetting}.
  In Table ~\ref{tab::imagenet-proj} we follow the experimental setting in ~\cite{Guo18inputtrans} and ~\cite{Dezfooli18patch}.  
 Although we compare with other methods in their preferred resolution and datasets for a fair comparison, we note that one of the unique advantages of our method is that it performs well in various resolutions (in our experiments, from 32 to 224), while others can only work on a limited range of resolutions. 

Comparison results show that our method significantly improves the classification robustness against unknown black-box attacks and outperforms state-of-the-art methods in most types of attacks with a large margin. Moreover, our retrained model achieves high accuracy on clean data and is comparable to the clean model, which means we preserve rich fine details that allow the network to learn discriminative features. Furthermore, we also compare our method with the widely used adversarial training~\cite{Kurakin16ifgsm} and show that we achieve better results on unknown attacks (Appendix B).

\begin{table}[]
	\centering
	\caption{CIFAR-10 classification accuracy for adversarial examples on VGG-16 after defense by methods in comparison. All methods are trained and tested on their transformed data. ``Clean'' means accuracy of transformed clean data on each method. ``STL'' denotes STL transformation with a single universal set of filters. ``STL (cluster)'' denotes STL filters are chosen through latent space clustering. }
	\label{tab::cifar-proj}
	\scalebox{0.62}{
		\begin{tabular}{c|c|ccccc}
			\hline
			Defense & Clean & FGSM-0.08 & FGSM-0.04 & BIM   & DeepFool & CW \\
			\hline
			\hline
			No Defense & 0.9298 & 0.5816 & 0.6523 & 0.1803 & 0.1760 & 0.0936 \\
			\hline
			MagNet\cite{Meng17magnet} & \bf{0.9206} & 0.7393 & 0.8552 & 0.7707 & 0.8770& 0.8594 \\
			PixelDefend\cite{song18pixeldefend} & 0.9041 & 0.8316 & \bf{0.8799} & 0.8515 & 0.8827 & 0.8845 \\
			\hline
			STL        & 0.9002 & 0.8515 & 0.8732 & 0.8754 & 0.8838 &  0.8880\\
			STL (cluster)& 0.9011 & \bf{0.8567} & 0.8715 & \bf{0.8803} & \bf{0.8890} & \bf{0.8904}\\
			\hline
		\end{tabular}
	}
\end{table}

\begin{table}[]
	\centering
	\caption{ImageNet-10 classification accuracy for adversarial examples on VGG-16 after defense by methods in comparison at resolution 64 (Table ~\ref{tab::imagenet10-proj-64}) and 128 (Table ~\ref{tab::imagenet10-proj-128}). All methods are trained and tested on their transformed data by their defense method. Here Crop-Ens denotes Crop+TVM+Quilting in ~\cite{Guo18inputtrans} and PD-Ens denotes PD+R-CAM+DWT in ~\cite{Prakash18deflect}. }
	\begin{subtable}{0.5\textwidth}
	\caption{Resolution 64.}
	\scalebox{0.63}{
		\begin{tabular}{c|c|ccccc}
			\hline
			Defense & Clean & FGSM-0.08 & FGSM-0.04 & BIM   & DeepFool & CW \\
			\hline
			\hline
			No Defense  & 0.8665 & 0.2816& 0.3080 & 0.1883 & 0.0811 & 0.0751 \\
			\hline
			TVM\cite{Guo18inputtrans} & 0.7555 & 0.5997 & 0.6930 & 0.7156 & 0.7210 & 0.7187 \\
			Quilting\cite{Guo18inputtrans} & 0.7741 & 0.7304 & 0.7418 & 0.7642 & 0.7646 & 0.7662 \\
			Crop-Ens\cite{Guo18inputtrans} & 0.7508 & 0.6968 & 0.7221 & 0.7369 & 0.7401 & 0.7304 \\
			PD-Ens\cite{Prakash18deflect} & 0.8250 & 0.6634 & 0.7607 & 0.7903 & 0.7955 & 0.7813 \\
			\hline
			STL   & \bf{0.8438} & 0.7275 & 0.8002 & \bf{0.8164} & 0.8163 & 0.8058 \\
			STL (cluster)& 0.8421 & \bf{0.7514} & \bf{0.8038} & 0.8103 & \bf{0.8221} & \bf{0.8122} \\
			\hline
		\end{tabular} 
	}
	\label{tab::imagenet10-proj-64}
	\end{subtable}
	\begin{subtable}{0.5\textwidth}
	\caption{Resolution 128.}
	\scalebox{0.63}{
	\label{tab::imagenet10-proj-128}
		\begin{tabular}{c|c|ccccc}
			\hline
			Defense & Clean & FGSM-0.08 & FGSM-0.04 & BIM   & DeepFool & CW \\
			\hline 
			\hline
			No Defense & 0.8991 & 0.2123 & 0.2409 & 0.1790 & 0.0584 & 0.0504 \\
			\hline
			TVM\cite{Guo18inputtrans} & 0.8567 & 0.7302 & 0.8181 &0.8183&0.8221 &0.8101  \\
			Quilting\cite{Guo18inputtrans} & 0.8354 & \bf{0.7612} & 0.7914 & 0.8048 & 0.8164 & 0.8093 \\
			Crop-Ens\cite{Guo18inputtrans} & 0.8382 & 0.7640 & 0.7969 & 0.8033 & 0.8071 & 0.7955 \\
			PD-Ens\cite{Prakash18deflect} & 0.8603 & 0.6740 & 0.8011 & 0.8273 & 0.8320 & 0.8262 \\
			\hline
			STL   & \bf{0.8784} & 0.7202 & 0.8308 & 0.8320 & \bf{0.8560} & \bf{0.8449} \\
			STL (cluster) & 0.8721 & 0.7421 & \bf{0.8356} &\bf{0.8385} & 0.8494 & 0.8421 \\
			\hline
		\end{tabular} 
	}
	\end{subtable}
	\label{tab::imagenet10-proj}
\end{table}

\begin{table}[]
	\centering
	\caption{Top-1 ImagetNet classification accuracy for adversarial examples on ResNet-50 after defense by methods in comparison. 
  We follow experimental settings in \cite{Guo18inputtrans} and ~\cite{Dezfooli18patch} where all attacks are in an average normalized $L_2$-dissimilarity of 0.06. All methods are trained and tested on their transformed data.}
	\label{tab::imagenet-proj}
	\scalebox{0.65}{
		\begin{tabular}{c|c|ccccc}
			\hline
			Defense & Clean & FGSM &  BIM   & DeepFool & CW &UAP\\
			\hline
			\hline
			No Defense & 0.761 & 0.107 & 0.012 & 0.010 & 0.019 & 0.133 \\
			\hline 
			quilt\cite{Guo18inputtrans} & 0.701 & 0.655 & 0.656 & 0.652 & 0.641 & -\\
			TVM+quilt\cite{Guo18inputtrans} & \bf{0.724} & 0.657 & 0.658 & 0.658 & 0.640 & -\\
			Crop-Ens\cite{Guo18inputtrans} & 0.721 & 0.667 & 0.670 & 0.671 & 0.635& - \\
			D3 (40K-5)\cite{Dezfooli18patch} & 0.718 & 0.686 & - & 0.631 & - & \bf{0.715} \\
			D3 (10K-5)\cite{Dezfooli18patch} & 0.708 & 0.683 & - & 0.646 & - & 0.703 \\
			D3 (10K-4)\cite{Dezfooli18patch} & 0.690 & 0.671 & - & 0.648 & - & 0.689\\
			PD-Ens\cite{Prakash18deflect} & 0.719 & 0.637 & 0.633 & 0.638 & 0.643 & 0.667 \\
			\hline
			STL (cluster) & 0.721 & \bf{0.693} & \bf{0.678} & \bf{0.685} & \bf{0.677} & 0.712\\
			\hline
		\end{tabular} 
	}
\end{table}

\subsection{White Box Attacks} 
\begin{table}[]
	\centering
	\caption{Backward Pass Differentiable Approximation (BPDA)~\cite{Athalye18obfuscated} attack results on CIFAR-10, VGG-16. All methods are attacked at distance $L_\infty=0.031$. Defenses denoted with $\ast$ propose combining adversarial training. }
	\label{tab::obfuscated-cifar}	
	\scalebox{0.61}{
		\begin{tabular}{c|ccccc|cc}
			\hline
			Defense  & SAP~\cite{Dhillon2018sap} & TE~\cite{buckman2018thermometer} & LID~\cite{ma2018lid} & PD~\cite{song18pixeldefend} & MagNet~\cite{Meng17magnet} & STL & STL (cluster)\\
			\hline
			Accuracy & 0.00 & 0.00* & 0.05 & 0.09* & 0.10 & 0.38* & \bf{0.42*} \\
			\hline
		\end{tabular} 
	}
\end{table}

Our defense is designed primarily for black/grey-box attacks, and like other methods, is highly susceptible to white-box attacks, especially on ImageNet~\cite{Deng09imagenet}. Nevertheless, we show that our method is significantly  less susceptible to the white-box attack Backward Pass Differentiable Approximation (BPDA) on CIFAR-10~\cite{Athalye18obfuscated}. 
BPDA specifically targets defenses in which the gradient does not optimize the loss; this is the case for our method since our STL is non-differentiable.
Table~\ref{tab::obfuscated-cifar} shows that although our defense accuracy is hurt by obfuscated gradient-based attacks, it is much more robust than other defenses with this phenomenon on CIFAR-10 dataset. 

On ImageNet~\cite{Deng09imagenet}, all defense methods in their case study (\cite{Guo18inputtrans} and \cite{xie2018mitigating}) get $0\%$ defense accuracy. Under the same settings, our defense accuracy similarly collapses to $1\%$. We further analyze our method's robustness to other simple custom-made white-box attacks with full knowledge of our model (including dictionary coefficients) in Appendix C.


\subsection{Input Transformation Effectiveness}
Since STL has a strong reconstruction capacity, the projected images still faithfully preserve information from the input data space. This is a useful property since it allows us to use a vanilla model to partially defend against adversarial examples when we are not able train our own classifier on quasi-natural images due to limitations such as access to the entire dataset.

Hence, we also evaluate the accuracy of using STL to project adversarial examples of a vanilla model that was pre-trained only on clean data. 
To perform the defense, we simply project the input into quasi-natural space and feed the projected image back into the vanilla model.  

\begin{figure}
	\centering
	\includegraphics[width=1\linewidth]{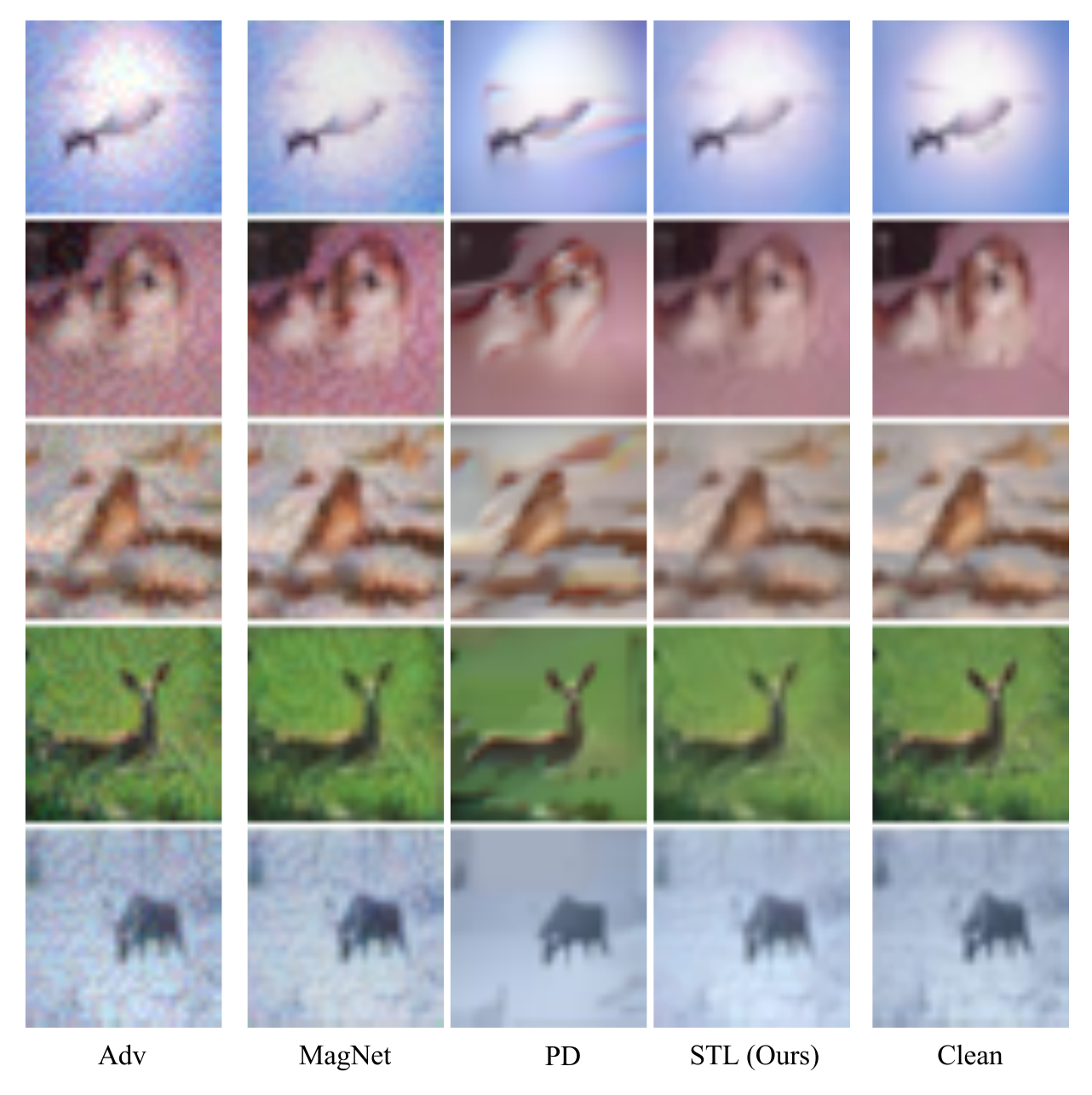}
	\caption{Qualitative comparison on CIFAR-10~\cite{cifar} with MagNet~\cite{Meng17magnet} and PixelDefend (PD)~\cite{song18pixeldefend}. 
  The first column is adversarial images generated by FGSM~\cite{Goodfellow14fgsm} with $L_2$-dissimilarity $=$ 0.08.
  The last column is corresponding clean images. We can observe that MagNet cannot fully remove adversarial perturbation, while PixelDefend over-smooths images, causing large information loss and sometimes introducing colorful artifacts. }
	\label{fig::comparison_cifar}
\end{figure}

We compare with other input-transformation methods applied to attacked vanilla models in Table ~\ref{tab::cifar-clean} for CIFAR-10, Table ~\ref{tab::imagenet10-clean} for ImageNet-10 and Table ~\ref{tab::imagenet-clean}). 
 Qualitative comparisons of our input transformations are shown in Figure ~\ref{fig::comparison_cifar} for CIFAR-10 and Figure ~\ref{fig::comparison_imagenet} for ImageNet. More results are in Appendix E. 

Under relatively large perturbations (e.g. FGSM-0.08), all competing methods fail to successfully overcome adversarial attacks while our method significantly outperforms them. On slightly perturbed adversarial examples (e.g. DeepFool and CW), we achieve a strong defense and also maintain accuracy on clean data. 
We see that our method can effectively defend against adversarial attacks even using a vanilla clean model.

\begin{table}[]
	\centering
	\caption{CIFAR-10 classification accuracy of  transformed clean and adversarial examples on the attacked vanilla VGG-16 model.}
	\label{tab::cifar-clean}
	\scalebox{0.65}{
		\begin{tabular}{c|c|ccccc}
			\hline
			Defense & Clean & FGSM-0.08 & FGSM-0.04 & BIM   & DeepFool & CW \\
			\hline
			\hline
			No Defense& 0.9298 & 0.5816 & 0.6523 & 0.1803 & 0.1760 & 0.0936 \\
			\hline
			MagNet\cite{Meng17magnet}& \bf{0.9035} & 0.6145 & 0.6521 & 0.4312 & 0.6535 & 0.4845 \\
			PixelDefend\cite{song18pixeldefend}& 0.8526 & 0.6810 & 0.7329 & \bf{0.7729}  &0.7414  & 0.7579  \\
			\hline
			STL& 0.8285 & 0.7099 & 0.7487 & 0.7462 & 0.7854 & 0.7765 \\
			STL (cluster)  & 0.8360 & \bf{0.7103} & \bf{0.7547} & 0.7531 & \bf{0.7959} & \bf{0.7906}\\
			\hline
		\end{tabular}
	}
\end{table}

\begin{table}[]
	\centering
	\caption{ImageNet-10 classification accuracy of  transformed clean and adversarial examples on an attacked vanilla VGG-16 model at resolution 64 (Table~\ref{tab::imagenet10-clean-64}) and 128 (Table~\ref{tab::imagenet10-clean-128}).}
	\begin{subtable}{0.5\textwidth}
	\caption{Resolution 64.}
	\scalebox{0.63}{
		\begin{tabular}{c|c|ccccc}
			\hline
			Defense & Clean & FGSM-0.08 & FGSM-0.04 & BIM   & DeepFool & CW \\
			\hline
			\hline
			No Defense& 0.8665 & 0.2816 & 0.3080 & 0.1883 & 0.0811 & 0.0751 \\
			\hline
			TVM\cite{Guo18inputtrans} & \bf{0.8172} & 0.3403 & 0.4744 & 0.6595 & 0.6943 & 0.6823 \\
			Quilting\cite{Guo18inputtrans} & 0.6318 & 0.4541 & 0.5312 & 0.5696 & 0.5436 & 0.5563 \\
			Crop-Ens\cite{Guo18inputtrans} & 0.5590 & 0.4570 & 0.5328 & 0.5369  & 0.5429 & 0.5320 \\
			PD-Ens\cite{Prakash18deflect}  & 0.7946 & 0.3388 &  0.5526 & 0.6568 & 0.6919 & 0.6827 \\
			\hline
			STL       &0.7925 & 0.5472 & 0.6825 & \bf{0.7245} & 0.7562 & 0.7414 \\
			STL (cluster)&0.8017 & \bf{0.5729} & \bf{0.6914} & 0.7234 & \bf{0.7652} & \bf{0.7521} \\
			\hline
		\end{tabular} 
	}
	\label{tab::imagenet10-clean-64}
	\end{subtable}
	\begin{subtable}{0.5\textwidth}
	\caption{Resolution 128.}
	\label{tab::imagenet10-clean-128}
	\scalebox{0.63}{
		\begin{tabular}{c|c|ccccc}
			\hline
			Defense & Clean & FGSM-0.08 & FGSM-0.04 & BIM   & DeepFool & CW \\
			\hline
			\hline
			No Defense & 0.8991 & 0.2123 & 0.2409 & 0.1790 & 0.0584 & 0.0504 \\
			\hline
			TVM\cite{Guo18inputtrans}& 0.8591 & 0.2568 & 0.4386 & 0.6586 & 0.6360 & 0.6129 \\
			Quilting\cite{Guo18inputtrans}& 0.8149 & 0.3903 & 0.5889 & 0.6434 & 0.6242 & 0.5922 \\
			Crop-Ens\cite{Guo18inputtrans} & 0.7730 & 0.4622 & 0.6447 & 0.6876 & 0.7060 & 0.6888 \\
			PD-Ens\cite{Prakash18deflect} & \bf{0.8789} & 0.2333 & 0.4286 & 0.7221 & \bf{0.7359} & 0.7272 \\
			\hline
			STL         &0.8654 & 0.4552 & 0.6418 & \bf{0.7332} & 0.7308 & 0.7212 \\
			STL (cluster)&0.8759 & \bf{0.4733} & \bf{0.6606} & 0.7323 & 0.7301 & \bf{0.7432} \\
			\hline
		\end{tabular} 
	}
	\end{subtable}
	\label{tab::imagenet10-clean}
\end{table}

\begin{table}[]
	\centering
	\caption{Top-1 ImageNet classification accuracy of  transformed clean and adversarial examples on an attacked vanilla ResNet-50 model.}
	\label{tab::imagenet-clean}
	\scalebox{0.63}{
		\begin{tabular}{c|c|ccccc}
			\hline
			Defense & Clean & FGSM-0.08 & FGSM-0.04 & BIM   & DeepFool & CW \\
			\hline
			\hline
			No Defense & 0.7613 & 0.0862 & 0.1140 & 0.0131 & 0.0106 & 0.0201\\
			\hline
			TVM\cite{Guo18inputtrans} & 0.6205 & 0.3123 & 0.4256 & 0.4923 & 0.5232 & 0.5012 \\
			Quilting\cite{Guo18inputtrans} & 0.4168 & 0.3787 & 0.3865 & 0.3823 & 0.3859 & 0.3783 \\
			Crop-Ens\cite{Guo18inputtrans} & 0.6432 & 0.4623 & 0.5546 & 0.5965 & 0.6023 & 0.5980 \\
			PD-Ens\cite{Prakash18deflect} & 0.6821 & 0.3846 & 0.5691 & 0.6089 & 0.6220 & \bf{0.6371} \\
			\hline
			STL         &0.6728 & 0.5348 & 0.6032 & 0.6253 & 0.6233 & 0.6158 \\
			STL (cluster)&\bf{0.6921} & \bf{0.5588} &\bf{0.6053} & \bf{0.6348} & \bf{0.6468} & 0.6220 \\
			\hline
		\end{tabular} 
	}
\end{table}

\subsection{Trade-off Between Quality and Robustness}

In transformation-based adversarial defenses, we typically aim to remove adversarial perturbations while preserving useful details. However, this is hard to achieve, as important details and adversarial perturbations are usually removed together.
Thus, we examine the inherent trade-off between transformation quality and defensive robustness in our method. 
  
\begin{figure}
	\centering
	\includegraphics[width=0.9\linewidth]{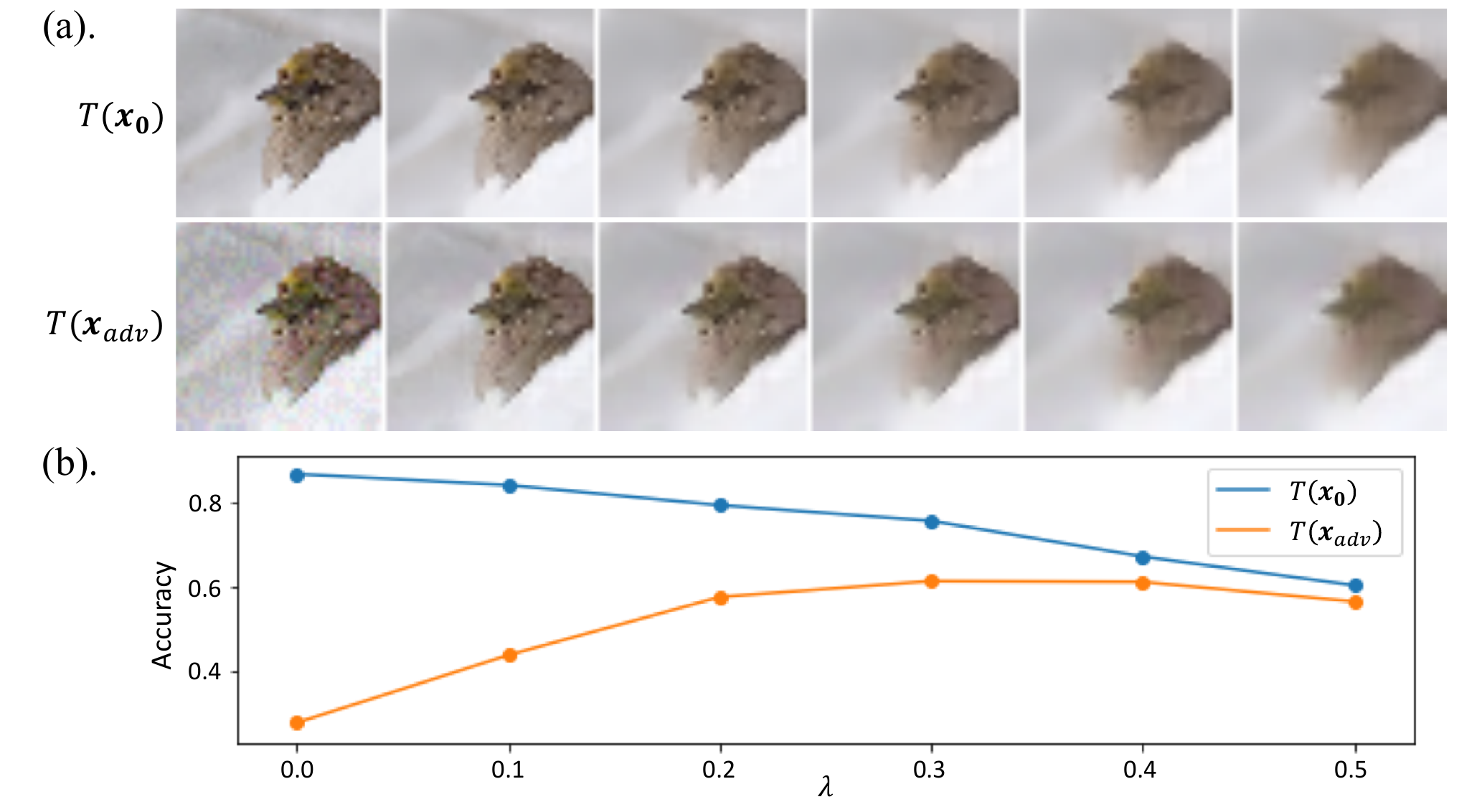}
	\caption{Intrinsic tradeoff between image reconstruction quality and defensive robustness.  (a). Transformation results of each corresponding $\lambda$.  (b). Accuracy of $T (\bm{x}_{adv}$ and $T (\bm{x_0})$ on attacked vanilla model. (Setting: FGSM-0.08, ImageNet-10, VGG-16, resolution 64).}
		\vspace{-3mm}
	\label{fig::tradeoff}
\end{figure}

In our method, the key parameter controlling the projection quality is the sparsity constraint weight $\lambda$: a larger $\lambda$ causes more blurry results. We gradually increase $\lambda$ and explore this trade-off (Figure ~\ref{fig::tradeoff}). We denote $\mathrm{Acc} (\bm{x})$ as the accuracy on the vanilla model of input $\bm{x}$. Higher $\mathrm{Acc} (T (\bm{x_0}))$ means higher transformation quality because the projected images still preserve useful information. Small $\|\mathrm{Acc} (T (\bm{x_0}))-\mathrm{Acc} (T (\bm{x}_{adv}))\|$ means the clean and adversarial examples are similar in feature space. The decision can be robust if both $\mathrm{Acc} (T (\bm{x_0}))$ and $\mathrm{Acc} (T (\bm{x}_{adv}))$ are high.
In Figure ~\ref{fig::tradeoff}, we see that as $\lambda$ increases, $\mathrm{Acc} (T (\bm{x_0}))$ decreases and the gap between $\mathrm{Acc} (T (\bm{x_0}))$ and $\mathrm{Acc} (T (\bm{x}_{adv}))$ shrinks. 

\begin{figure}
	\centering
	\vspace{-2mm}
	\includegraphics[width=0.7\linewidth]{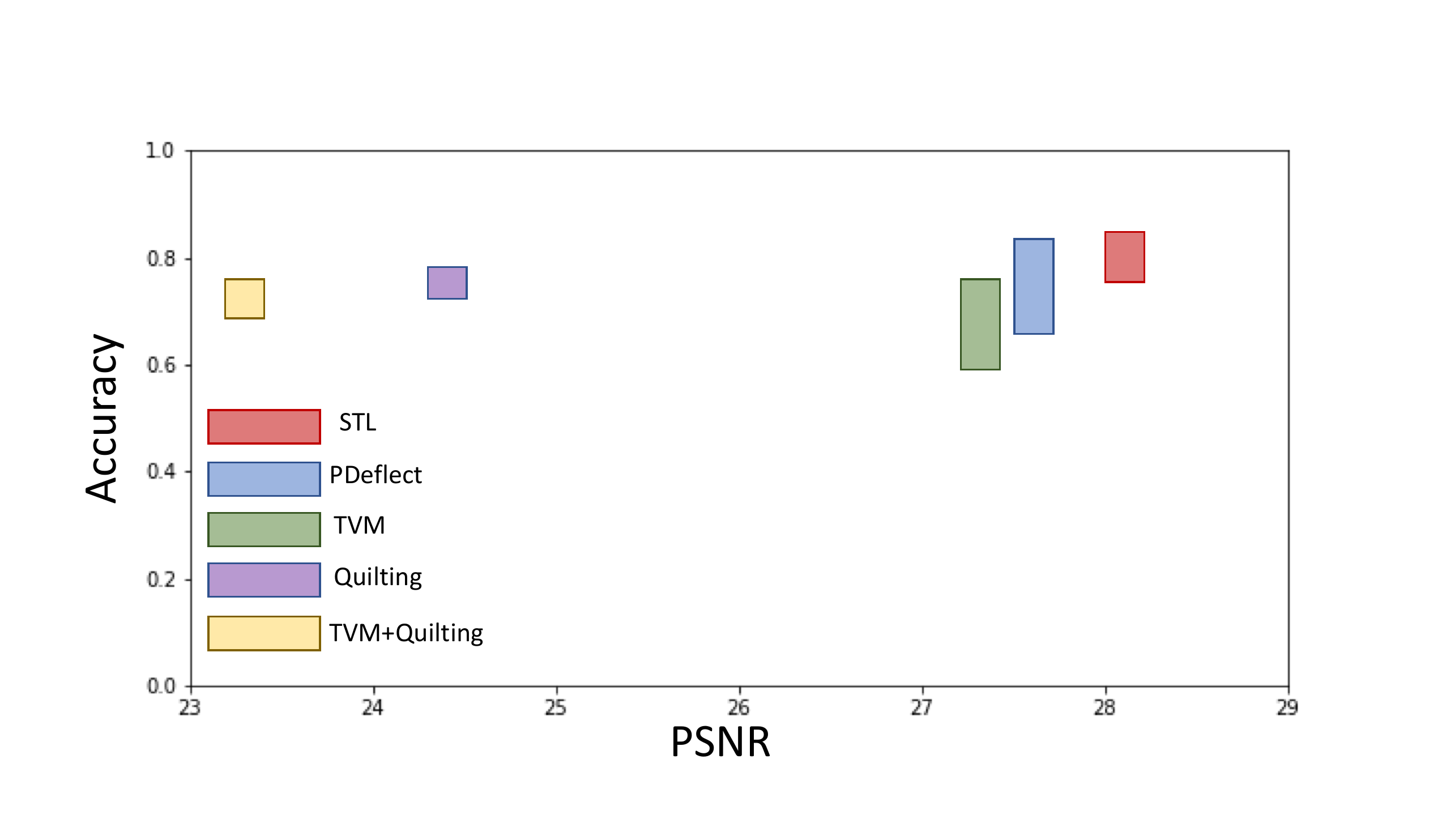}
	\vspace{-4mm}
	\caption{The PSNR, $\mathrm{Acc} (T (\bm{x}_{adv}))$ and $\mathrm{Acc} (T (\bm{x_0}))$ of different methods (Setting: FGSM-0.08, ImageNet-10, VGG-16, resolution 64). For both axes, the higher number the better. And less difference between $\mathrm{Acc} (T (\bm{x}_{adv}))$ and $\mathrm{Acc} (T (\bm{x_0}))$ means higher robustness.}
	\label{fig::psnr_acc}
\end{figure}

We additionally propose a metric to measure this tradeoff. Specifically, we use PSNR between $T (\bm{x}_{adv})$ and $\bm{x_0}$ to measure reconstruction quality. For each method in comparison, let $a_0=\mathrm{Acc} (T (\bm{x_0}))$ and $a_{adv}=\mathrm{Acc} (T (\bm{x}_{adv}))$, then we associate it with a characteristic interval $[\min(a_0,a_{adv}), \max(a_0, a_{adv})]$ to represent its overall prediction quality. Apparaently, a strong method should have an interval that is short (good robustness) and high (good accuracy). We plot a 2D PSNR vs. prediction quality map, where the top right corner indicates highest robustness and prediction quality. In Figure ~\ref{fig::psnr_acc}, we show comparison results of occupied regions on this map. Our method achieves both the highest PSNR and most preferable characteristic interval, demonstrating its superior ability to achieve robustness, accuracy, and maintain image quality. 

\section{Conclusion}
We have proposed a novel state-of-the-art attack-agnostic adversarial defense method with additional increased robustness to input resolution, perturbation scale, and dataset scale. Inspired by convolutional sparse coding, we design a novel sparse transformation layer (STL) to project the inputs to a low-dimensional quasi-natural space, wherein a retrained classifier can make more reliable decisions. We evaluate the proposed method on CIFAR-10 and ImageNet and show that our defense mechanism provide state-of-the-art results. We have also provided an analysis of the trade-off between the  projection image quality and defense robustness.
\vspace{-2mm}
\paragraph{Acknowledgements} We thank Bo Li for providing key discussions on white box attacks.

{\small
\bibliographystyle{ieee}
\bibliography{egbib}
}

\begin{appendices}
 \clearpage

\section{Overview}
In addition to the experiments shown in the main paper, we also compare our method with adversarial training to emphasize the generalizability of our method to different attack types (Section~\ref{app::comparion_to_adversarial_training}). We also test our performance under some custom-made white-box attacks (Section~\ref{app::white_box}). 
We then analyze several key parameters of our algorithm from the aspects of robustness and reconstruction quality (Section~\ref{app::parameter_analysis}). 
Lastly, we provide visualizations demonstrating the superiority of STL in achieving state-of-the-art performance without loss of image quality (Section~\ref{app::more_results}). 
 \vspace{-1mm}
\section{More Comparisons}

\label{app::comparion_to_adversarial_training}
Through adversarial training, the model can reach high robustness in defending a designated attack, but still has poor performance to unknown attacks.

In Table~\ref{tab::adv_training_64}, we compare our method with networks adversarially trained~\cite{Goodfellow14fgsm,Kurakin16ifgsm,tramer2018ensemble} on a designated attack method (FGSM attack). Although our method performs slightly worse than adversarial training using data generated from the already-known attack method, we do achieve comparable, sometimes even better performance, on novel unknown attacks. Please read the caption of Table 1 for more details.

 \vspace{-1mm}
\begin{table}[H]
	\centering
	\caption{Comparison with adversarial training. Attacks are named by type-$L_2$ dissimilarity. Adversarial training was performed on FGSM-0.08 following the popular method introduced in ~\cite{Kurakin16ifgsm}. On the designated attack method (FGSM attacks with other parameters), our method performans slightly worse than adversarially trained version, but significantly better on the Uni attack (Universal perturbation~\cite{Dezfooli17universal}), which is an unknown attack to the FGSM-based adversarial training.}
	\label{tab::adv_training}	
	\begin{subtable}{0.5\textwidth}
	\caption{CIFAR-10, VGG16.}
	\scalebox{0.61}{
		\begin{tabular}{c|c|cccc|c}
			\hline
			Defense & Clean & FGSM-0.04 & FGSM-0.08 & FGSM-0.12   & FGSM-0.20 & Uni-0.08 \\
			\hline
			\hline
			No Defense &0.9298 &0.6523 & 0.5816 & 0.3412 & 0.2002 & 0.6823 \\
			Adv Training& 0.9158 & 0.9075 & 0.8890 & 0.8558 & 0.7732 & 0.8282 \\
			STL(Cluster) &0.9011 & 0.8715 & 0.8567 & 0.8258& 0.7632& 0.8642\\
			\hline
		\end{tabular} 
	}
	  	\label{tab::adv_training_cifar}
   	\end{subtable}
    \begin{subtable}{0.5\textwidth}
   		\caption{ImageNet-10, VGG16, resolution 64.}
   	\scalebox{0.61}{
   	   		\begin{tabular}{c|c|cccc|c}
   			\hline
   			Defense & Clean & FGSM-0.04 & FGSM-0.08 & FGSM-0.12   & FGSM-0.20 & Uni-0.08 \\
   			\hline
   			\hline
   			No Defense &0.8665 & 0.3080 & 0.2816 & 0.2433 & 0.1887 &  0.3312\\
   			Adv Training& 0.8358 & 0.8260 & 0.7983 & 0.7520 & 0.6576 & 0.672\\
   			STL(Cluster) &0.8421 & 0.8038 & 0.7514 & 0.7021& 0.6468& 0.7721\\
   			\hline
   		\end{tabular} 
   		
   		}
   	\end{subtable}
   	   \label{tab::adv_training_64}
\end{table}

\section{White-box Attacks}
\label{app::white_box}

In Section 5.3, we have shown that although our method is extremely susceptible to white box attacks on ImageNet, we considerably beat all other methods on BPDA on CIFAR-10. In this section, we further analyze our defense under some other simple white-box attack settings.

The first attack leverages full knowledge of our dictionary and directly performs attacks in the quasi-natural image space. Under FGSM with $L_2 = 0.08$ on CIFAR-10, we achieved an accuracy of 0.6253 with our defense and an accuracy of 0.5021 when no defense was applied.
The second attack adversarially perturbs the sparse coefficients, which are then used to construct attacked images. Under the this attack setting, applying FGSM ($L_2 = 0.04$) on CIFAR-10, the classiﬁcation accuracy is reduced to 0.2515. We achieved 0.5628 defense accuracy by combining our defense with adversarial training. We see that simple white box attacks that use full knowledge of our dictionary are somewhat effective, but not as devastating as BPDA.


\section{Parameter Analysis}
\label{app::parameter_analysis}
\begin{figure}[H]
	\centering
	\includegraphics[width=0.75\linewidth]{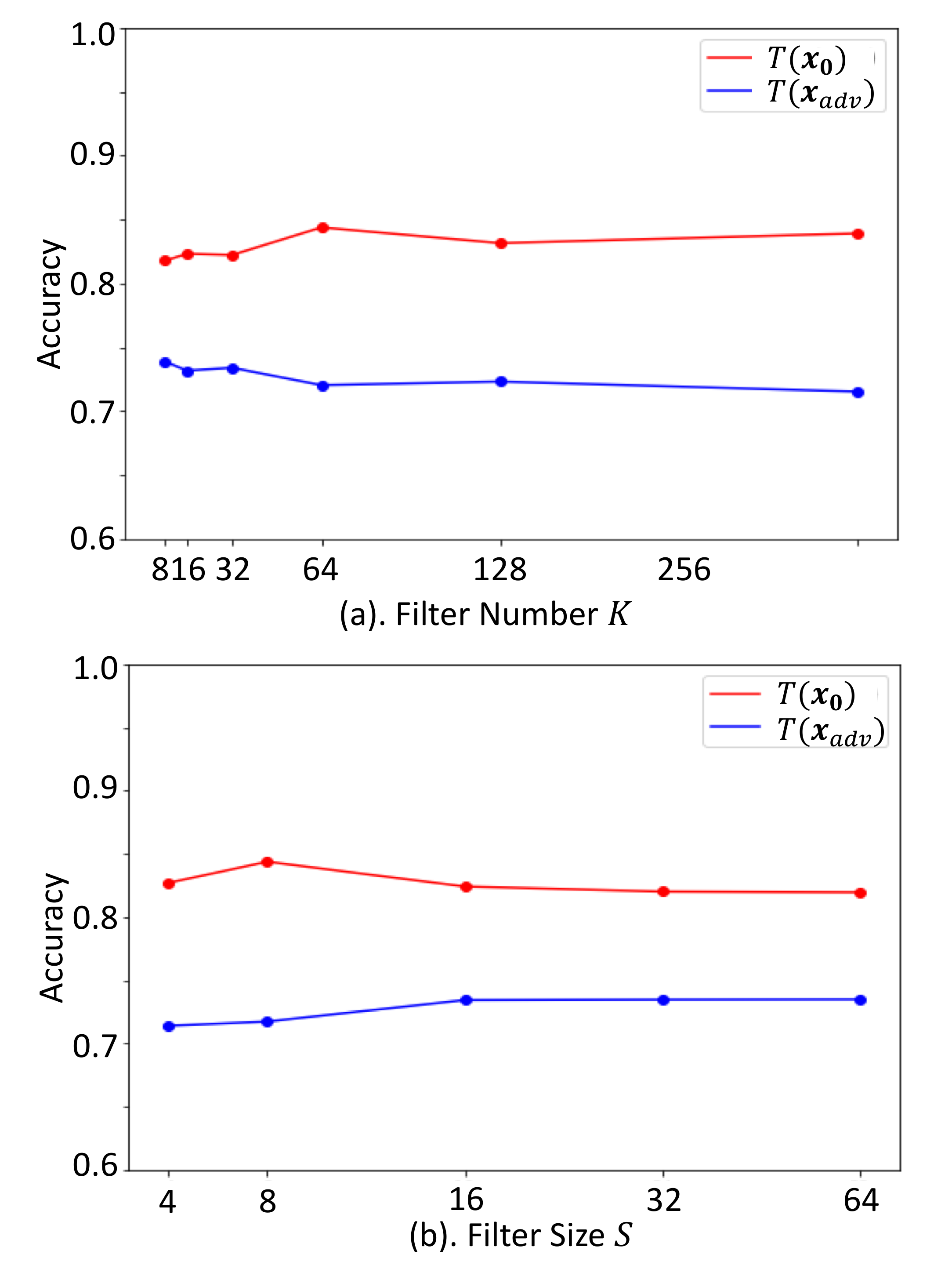}
	\caption{Parameter analysis of filter number $K$(a) and size $P$(b). Unless otherwise specified, the default setting is $K=64$, $S=8$, $\lambda=0.2$. All experiments are implemented on VGG16, ImageNet-10 at resolution 64.}
	\label{fig::parameter}
\end{figure}

\begin{figure*}
	\centering
	\includegraphics[width=0.9\linewidth]{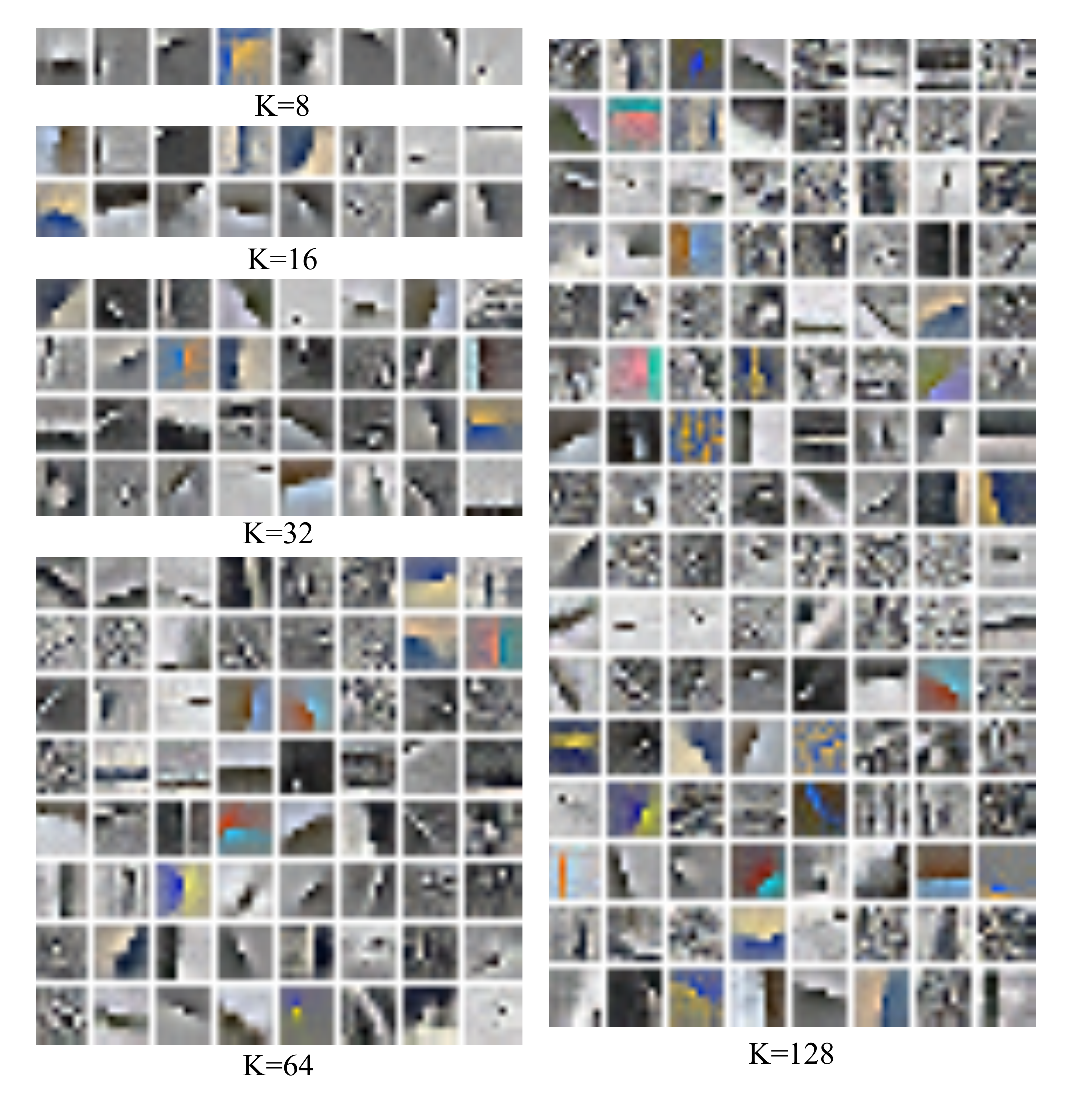}
	\caption{Filters of different number.}
	\label{fig::filter_k}
\end{figure*}

In Section 5.4 of the main paper, we have analyzed the impact of the sparsity regularization weight $\lambda$ in Eq~(2). In this section, we analyze the influence of other key hyper-parameters of our algorithm: filter number $K$, filter size $S$, and the number of subspace clusters $M$. We measure the prediction accuracy of the retrained model on transformed clean and adversarial data, denoted by $\mathrm{Acc} (T (\bm{x_0}))$ and $\mathrm{Acc} (T (\bm{x}_{adv})$, in Figure~\ref{fig::parameter}. The gap between the two numbers reflects the defensive robustness and the magnitude of each number reflects the reconstruction quality. 
 \vspace{-3mm}
\paragraph{Filter Number $K$}
As the  filter number $K$ increases, $\mathrm{Acc} (T (\bm{x_0}))$ also increases, because more filters naturally increases the representation power of the dictionary. However, on the other hand, $\mathrm{Acc} (T (\bm{x}_{adv}))$ decreases as a larger number of filters would inevitably introduce more components to characterize image details, hurting our method's ability to filter out unwanted adversarial perturbations. Performances w.r.t different number of filters are shown in Figure~\ref{fig::parameter} (a). The visualization of learned filters at different $K$s is shown in Figure~\ref{fig::filter_k}.

\vspace{-3mm}

\paragraph{Filter Size $S$}

\begin{figure*}
	\centering
	\includegraphics[width=0.9\linewidth]{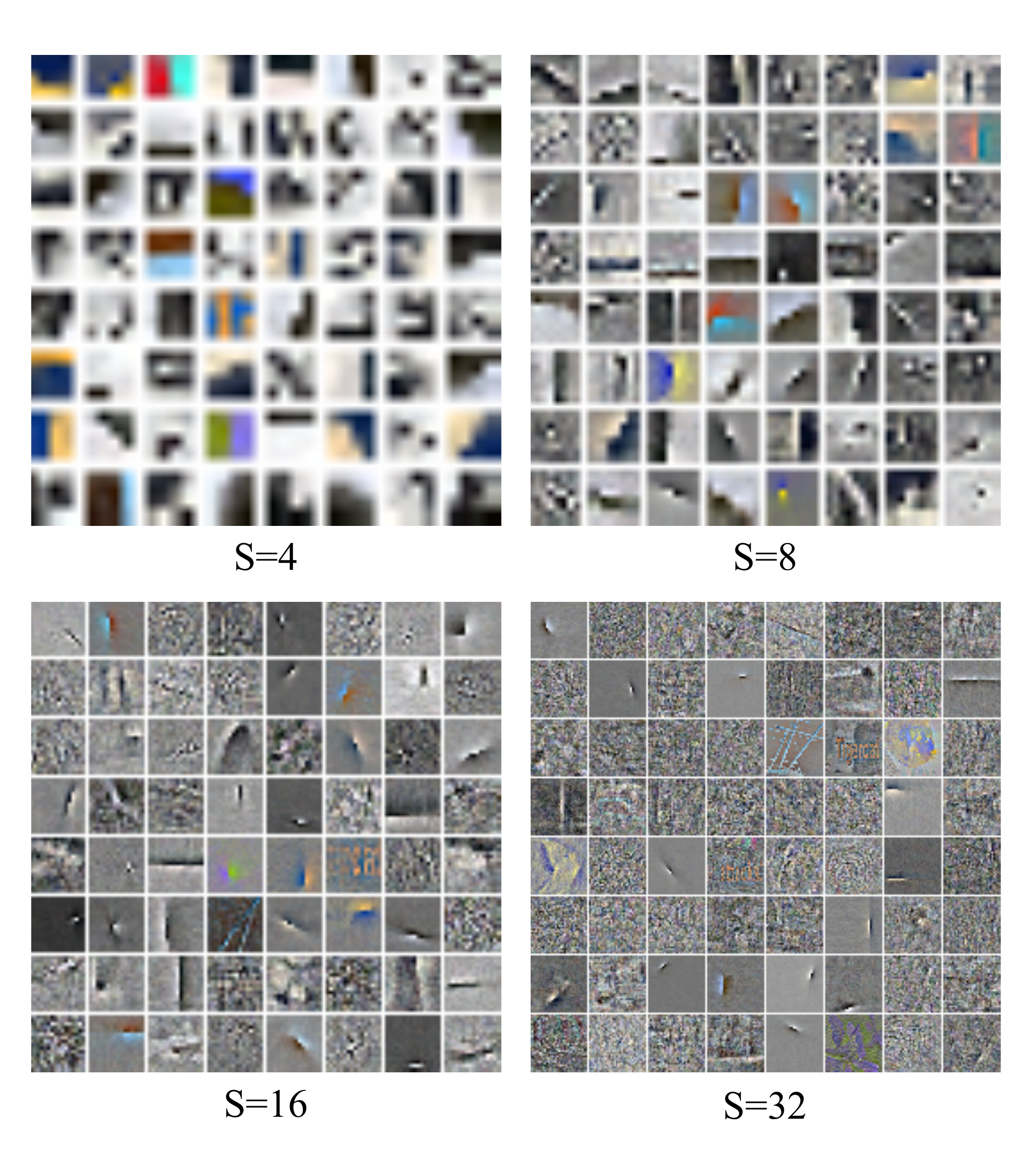}
	\caption{Visualization of filters of different sizes.}
	\label{fig::filter_p}
\end{figure*}

Figure~\ref{fig::parameter} (b) shows that our method is not sensitive to the selection of filter size. The visualization of filters with different sizes are shown in Figure~\ref{fig::filter_p}.

\begin{figure*}
	\centering
	\includegraphics[width=0.9\linewidth]{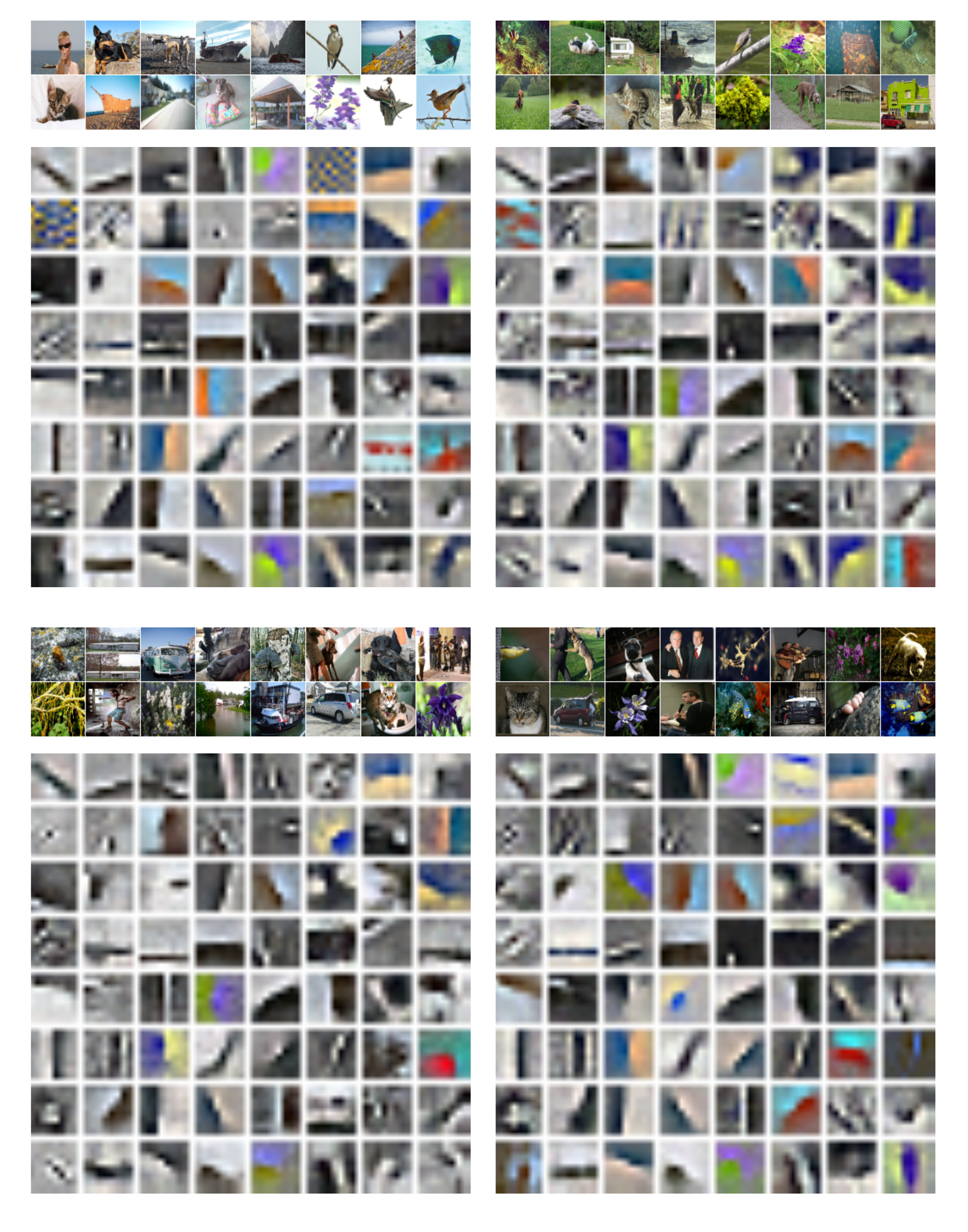}
	\caption{Filters and sample images for 4 clusters of ImageNet-10 at resolution 64.}
	\label{fig::cluster}
\end{figure*}

\vspace{-3mm}
\paragraph{Learned Filters for Individual Image Clusters}
In our experiments, we first split the natural data space into several clusters based on their DAE features and then learn individual dictionaries. 
The dictionary of each cluster will learn to capture some cluster-specific features. Filters and sample cluster images are shown in Figure ~\ref{fig::cluster}.  






 \vspace{-1mm}

\section{More Qualitative Results}
\label{app::more_results}
\begin{figure*}
	\centering
	\includegraphics[width=0.9\linewidth]{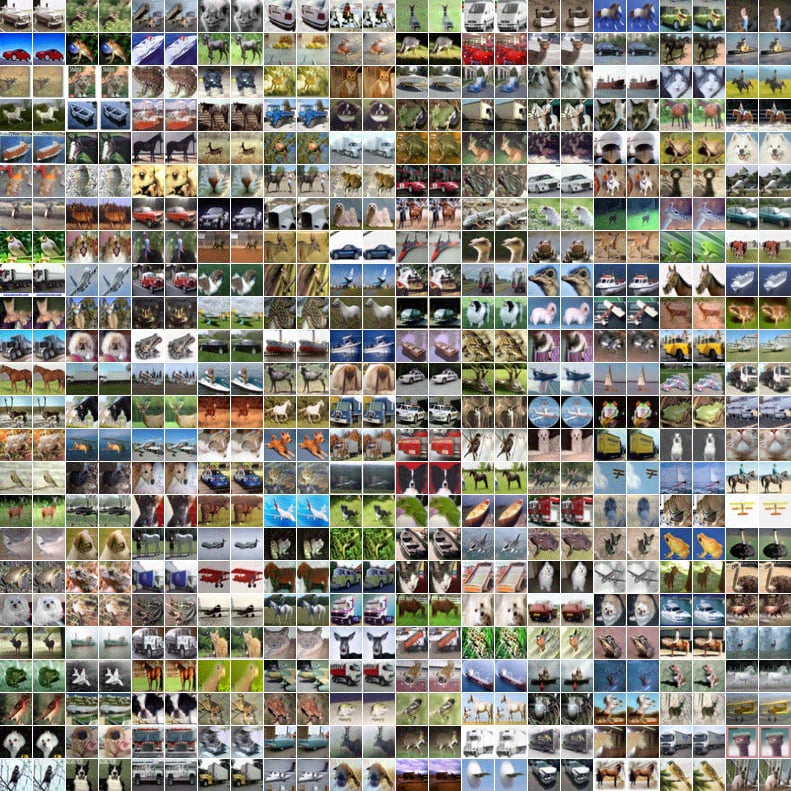}
	\caption{Transformation results for CIFAR-10. For every pair of images, the left is  the input adversarial image and the right is the transformed image. }
	\label{fig::more-cifar}
\end{figure*}

\begin{figure*}
	\centering
	\includegraphics[width=0.9\linewidth]{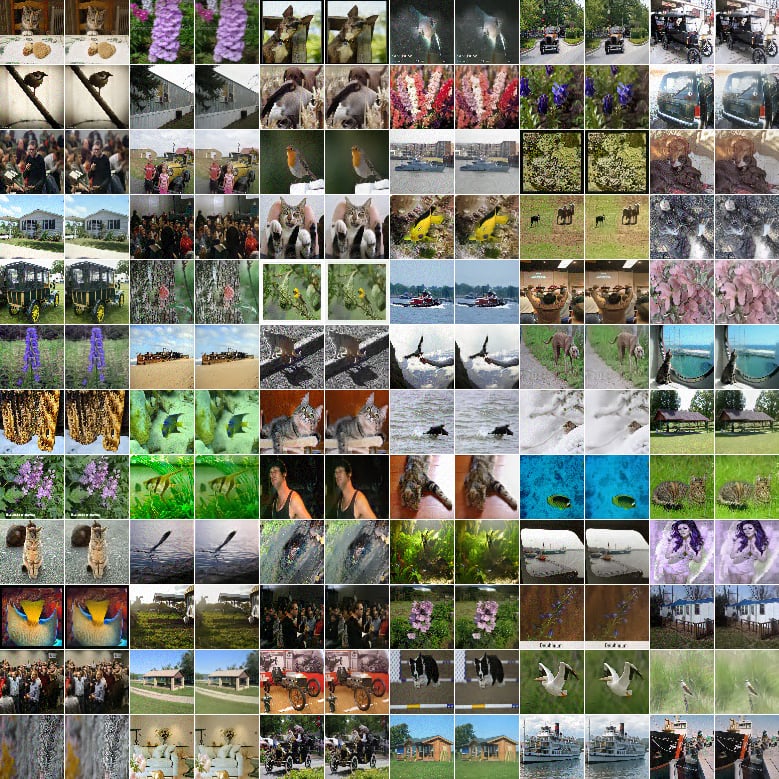}
	\caption{Transformation results for ImageNet-10 at resolution 64. For every pair of images, the left is the input adversarial image and the right is the transformed image. }
	\label{fig::more-64}
\end{figure*}

\begin{figure*}
	\centering
	\includegraphics[width=0.9\linewidth]{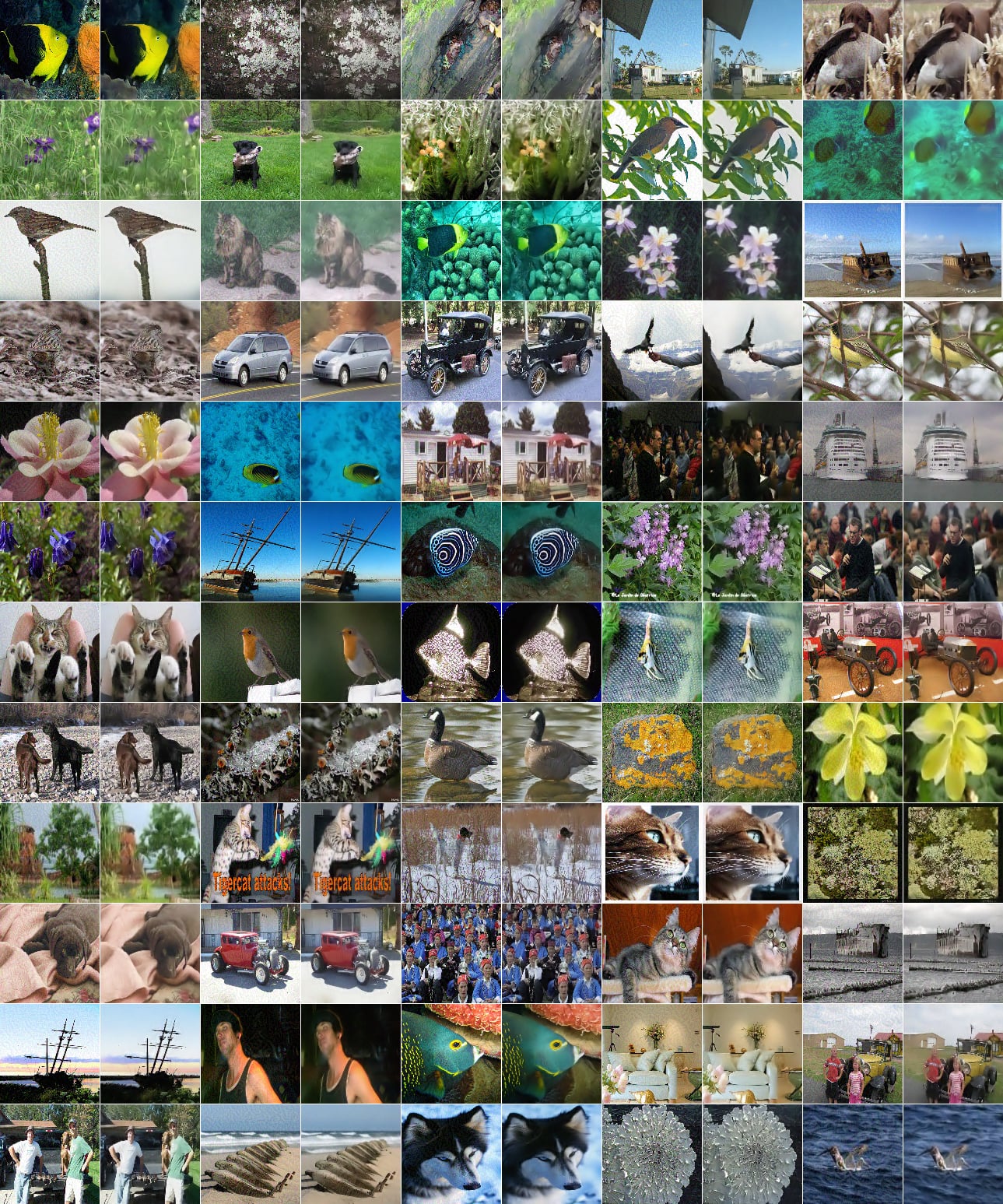}
	\caption{Transformation results for ImageNet-10 at resolution 128. For every pair of images, the left is the input adversarial image and the right is the transformed image.}
	\label{fig::more-128}
\end{figure*}

\begin{figure*}
	\centering
	\includegraphics[width=0.9\linewidth]{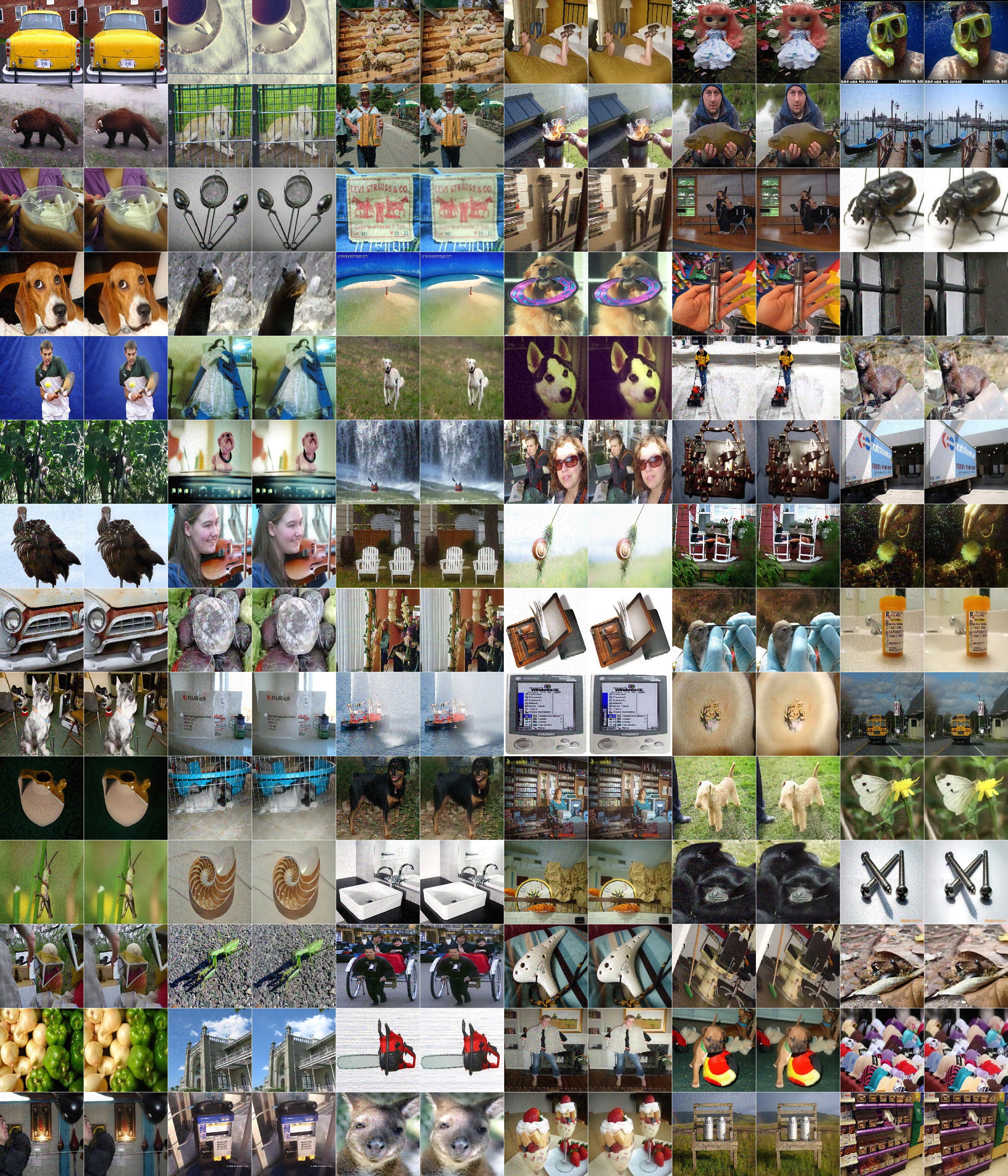}
	\caption{Transformation results for ImageNet-10 at resolution 224. For every pair of images, the left is the input adversarial image and the right is the transformed image. }
	\label{fig::more-224}
\end{figure*}

We show more transformation results on CIFAR-10~\cite{cifar} (Figure~\ref{fig::more-cifar}), ImageNet-10 (Figure.~\ref{fig::more-64} and Figure.~\ref{fig::more-128}), and ImageNet (Figure.~\ref{fig::more-224}).  


\end{appendices}



\end{document}